\documentclass[10pt,twocolumn,letterpaper]{article}

\usepackage{cvpr}
\usepackage{times}
\usepackage{epsfig}
\usepackage{graphicx}
\usepackage{amsmath}
\usepackage{amssymb}

\usepackage{booktabs}
\usepackage{indentfirst}

\usepackage[breaklinks=true,bookmarks=false]{hyperref}

\cvprfinalcopy 

\def\cvprPaperID{****} 

\begin{document}

\title{2nd Place and 2nd Place Solution to Kaggle Landmark Recognition and Retrieval Competition 2019}

\author{Kaibing Chen$^*$, Cheng Cui$^*$, Yuning Du$^*$, Xianglong Meng$^*$, Hui Ren\thanks{The authors contributed equally and they are ordered family alphabetically.}\\
Baidu Inc.\\
{\tt\small \{chenkaibing, v\_cuicheng, duyuning, mengxianglong01, renhui\}@baidu.com}
}

\maketitle

\begin{abstract}
We present a retrieval based system for landmark retrieval and recognition challenge.There are five parts in retrieval competition system, including feature extraction and matching to get candidates queue; database augmentation and query extension searching; reranking from recognition results and local feature matching. In recognition challenge including: landmark and non-landmark recognition, multiple recognition results voting and  reranking using combination of  recognition and retrieval results. All of models trained and predicted by PaddlePaddle framework\footnote{\url{https://github.com/PaddlePaddle/Paddle}}. Using our method, we achieved 2nd place in the Google Landmark Recognition 2019 and 2nd place in the Google Landmark Retrieval 2019 on kaggle. The source code is available at here\footnote{\url{https://github.com/PaddlePaddle/models/tree/develop/PaddleCV/Research/landmark}}.

\end{abstract}

\section{Introduction}

The Google Landmark Dataset(GLD) V2 is currently the largest publicly image retrieval and recogntion dataset\cite{noh2017large}, including 4M training data, more than 100,000 query images and nearly 1M index data. The large amounts of images in training dataset is the driving force of the generalizability of machine learning models. Successfully trained models on GLD V2 would push the frontier of image retrieval system with the help of data. 

The traditional image retrieval method employed invariant local feature\cite{lowe2004distinctive} incorporating with  spatial verification to achieve better results on many retrieval task with large scale or rotation transformations.  Generally,  the methods based on bag-of-word\cite{sivic2003video,csurka2004visual} are adopted on retrieval task , the alternative approach  likes Fisher vector\cite{sanchez2013image} and VLAD\cite{jegou2010aggregating} can generate compact feature which has better results on some task. 

With the development of deep learning,  more and more visual tasks use deep models to address their problems. In visual image retrieval,using embedding of deep convolutional networks can improve the performance significantly. We have trained deep convolutional backbones to extract  various image features. Our backbone model mainly uses ResNet152, ResNet200, SE\_ResNeXt152 and InceptionV4, which fine-tune the original network structure and add some training tricks by this paper\cite{xie2018bag}.The top1 validation accuracy of imagenet classification task is 80.61\%, 80.93\%, 81.45\%, 80.88\% respectively.

In this paper, we exploit retrieval based approach to solve landmark retrieval and recognition problem. We derive representation of image from trained multiple CNN models activations corresponding with PCA. The top candidates produced from NN search are feed into recognition and local feature matching module to get positive results. Then we re-rank the candidates with higher recognition and matching score, the following weighted DBA\cite{chum2007total} and QE\cite{chum2007total} are processed to generate final retrieval result. While due to recognition problem, we are focus on landmark and non-landmark detection and re-score with multi classification models instead of PCA, DBA and QE. 

\section{Retrieval method}

\begin{figure*}[ht!]
  \centering
  \includegraphics[width=0.9\linewidth]{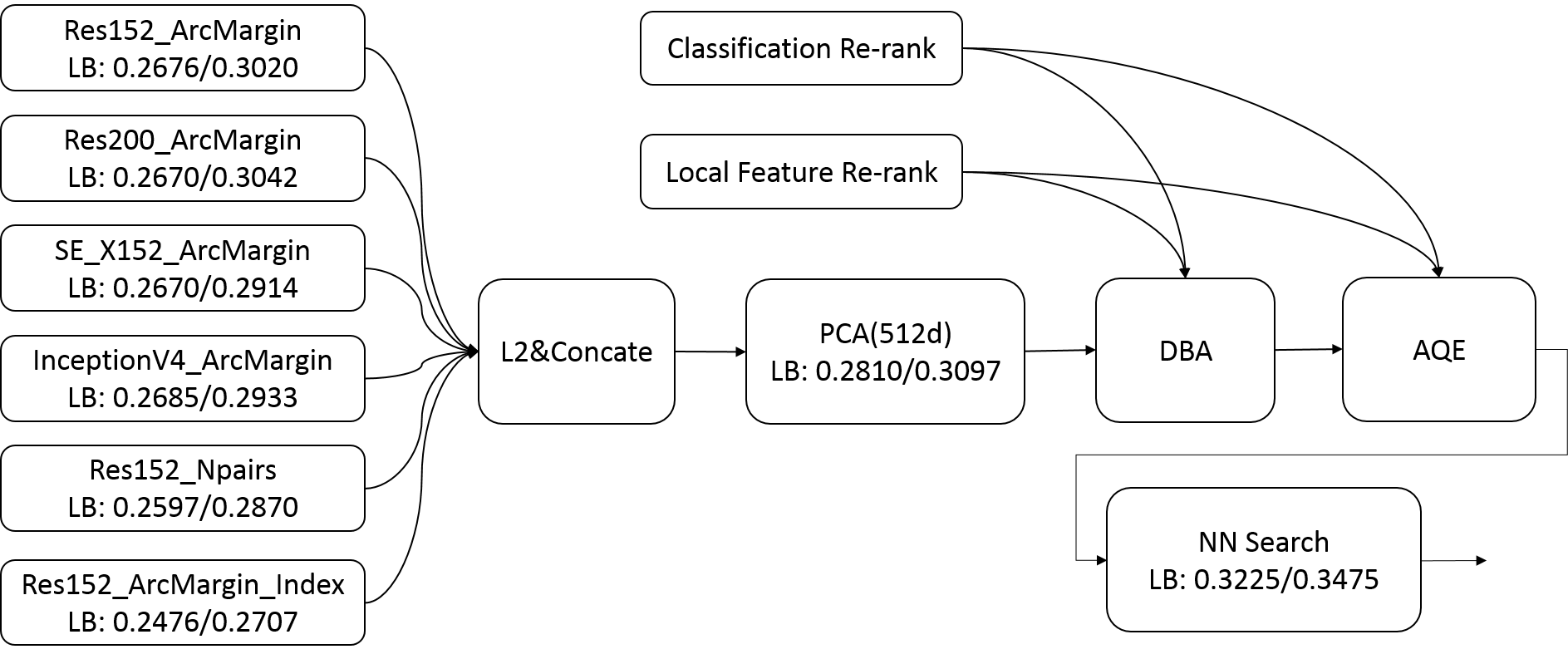}\\
  \caption{Our retrieval system ensemble six convolutional models, use pca to reduce the dimension of feature to 512 use DBA and AQE to re-rank our search results.}    
  \label{fig:retrieval system}
\end{figure*}

Our retrieval system is shown in Fig.~\ref{fig:retrieval system}. The detail of system is introduced later.

\subsection{Feature model}
\subsubsection{Global feature}

In this Kaggle retrieval competition, we fine-tune four convolutional neural networks on google landmark recognition 2019 dataset to extract our global image descriptors. The four convolutional backbones include ResNet152\cite{he2016deep}, ResNet200, SE\_ResNeXt152\cite{hu2018squeeze,xie2017aggregated} and InceptionV4\cite{szegedy2017inception}. Instead of using softmax loss for training, we train these models with arcmargin loss\cite{deng2018arcface}. Arcmargin loss is firstly employed in face recognition, we found that it can also produce distinguishing and compact descriptor in landmark dataset. In order to obtain compact descriptor from these backbone, we remove the last fully connected layer, and add two additional fully connected layers after average pooling layer. The output size of the first fully connected layer is 512, and the second is 203094 corresponding to the class number of training dataset. We finally select the output of the first fully connected layer as out image descriptor. We don't do any data cleaning when training models with arcmargin loss. We fine-tune models using SGD optimizer and keep training image size to 448. We believe that using large input size is benefit to extract feature of tiny landmark. 

We also use metric learning loss to learn image descriptor. Here we use ResNet152 backbone with Npairs loss\cite{sohn2016improved} to train. Similar to above setting, we also add a fully connected layer after average pooling layer to obtain image descriptor. Except for using google landmark recognition 2019 training dataset, we do clustering on stage2 index dataset to produce another training dataset. This dataset include 20w images and about 4w class number. We fine-tune this dataset using ResNet152 with arcmargin loss.  

For inference, we keep the short size of the image to 448, and feed the full image to the neural network. After obtaining above six descriptors, we normalize these descriptors and concatenate them together. In order to reduce the dimension of the descriptors, we use PCA\cite{jegou2012negative} trained on index dataset to reduce the dimension to 512. We use this descriptor and NN search to build retrieval system. 

All these backbones are pretrained on imagenet1k dataset using PaddlePaddle deep learning framework developed by Baidu. All the training methods had been open and the source code can be found here\footnote{\url{https://github.com/PaddlePaddle/models/tree/develop/PaddleCV/metric_learning}}.

\subsubsection{Local Feature}

\begin{figure}
  \includegraphics[width=0.95\linewidth]{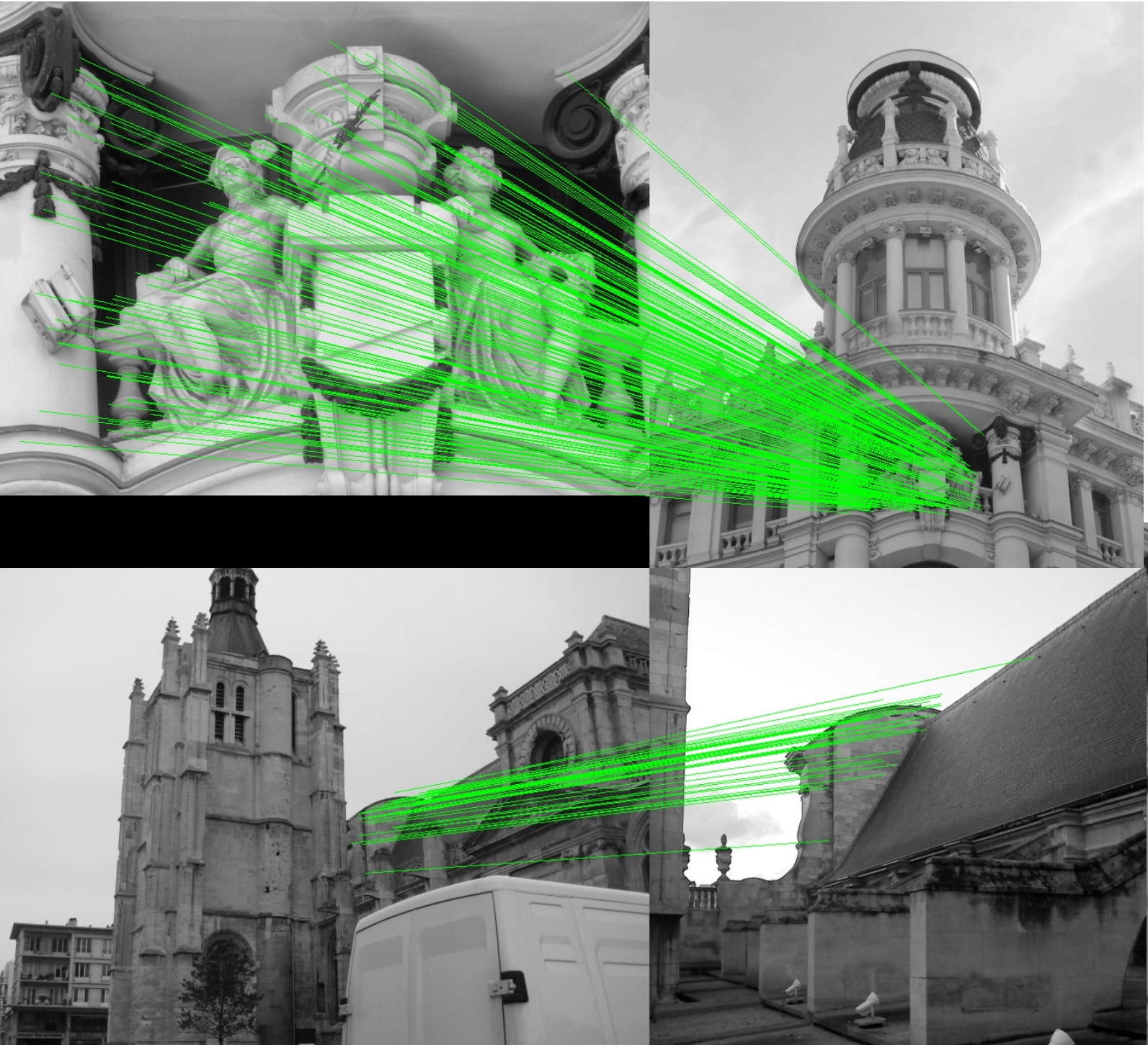}\\
  \caption{Difficult samples with local feature matching} 
  \label{fig:local match}
\end{figure}

Although CNN based local features\cite{zheng2017sift} has better accuracy in many datasets than traditional local features,the traditional local detector has stronger performance for scale and angle transformations. In kaggle retrieval datasets ,there are lots of samples with these transformations. Hard sample matching result show in Fig.~\ref{fig:local match}
 
According to the above problems, we select SURF\cite{bay2006surf} and Hassian-Affine\cite{medioni1987corner,mikolajczyk2005comparison} root sift\cite{arandjelovic2012three} as our local feature method. Our local feature image retrieval system is based on nearest neighbor search. To speed up the nearest neighbor search, we construct an inverted index which is implemented by a k-means clustering with 512 centers, for each point of query image we select top 20 clustering centers to search  top1 clustering center of all points descriptors in database set. 
 

\subsection{Recognition model}

In this kaggle retrieval competition, We use the full amount of data to train the classification model, including 4094044 images and 203094 classes.he main network used in the competition is ResNet152\cite{he2016deep} and InceptionV4\cite{szegedy2017inception}. In order to accelerate convergence and quickly verify the model effect, we add a embedding layer after ResNet C5, Which sizes is 256. Label smoothing\cite{xie2018bag} is used in the training model, and the soft label parameters is set to 0.1, 0.2. 

In addition to using the classification model for prediction, we also use the test and index data sets to retrieve the 4M train set with ResNet152 feature. If returned result category of top5 has only two categories and the max score retrieved is greater than 0.85, then the requested image is considered to be category with max voting number. We select the results of the same category of images from the index before all the search results. 

\subsection{Rank strategy}
As is known to us, query expansion (QE)\cite{chum2007total} and database augmentation (DBA) can improve performance of retrieval system significantly. Different to standard operation, we perform QE and DBA with classification re-rank and local feature re-rank. 

We perform database augmentation on the query and index dataset. We replace image descriptor with a weighted combination of itself and its top N neighbors. Specifically, we firstly use NN search to find top 300 neighbors for each images, then use classification model and local feature to verify on the top 300 neighbors to obtain M same landmark images. Finally we put the M same landmark images on the top of 300 neighbors. Precisely, we perform weighted sum-aggregation of the descriptors with weights computed using: 

\begin{equation}  
 N = \left\{
\begin{array}{rcl}
10               &  & {M \leq 10} \\
min(M, 20)       &  & {M > 10} \\
\end{array} \right.
\end{equation}

\begin{equation}
weights = \frac{N - x}{N}, 0 <= x < N
\end{equation}

For query Expansion, we use the same operation as database augmentation. We just modify the weights computed as: 

\begin{equation}  
 N = \left\{
\begin{array}{rcl}
3               &  & {M \leq 3} \\
min(M, 6)       &  & {M > 3} \\
\end{array} \right.
\end{equation}

\begin{equation}
weights = \frac{N - x}{N}, 0 <= x < N
\end{equation}

\subsection{Experiments}

The results of each model is listed here, we finally get public/private score is 0.3225 / 0.3475.

\begin{table}[ht!]
\caption{ Retrieval results of different models. }
  \centering
  \begin{tabular}{l|c c}
  \toprule
    model & public & private \\
    \midrule
    res152\_arcmargin & 0.2676 & 0.3020 \\
    res200\_arcmargin & 0.2670 & 0.3042 \\
    se\_x152\_arcmargin & 0.2670 & 0.2914 \\
    inceptionv4\_arcmargin & 0.2685 & 0.2933 \\
    res152\_npairs & 0.2597 & 0.2870 \\
    res152\_arcmargin\_index & 0.2476 & 0.2707 \\
    \midrule
    concate\_pca & 0.2810 & 0.3097 \\
    concate\_pca\_dba\_aqe & 0.3096 & 0.3345 \\
    concate\_pca\_dba\_aqe\_rerank & 0.3225 & 0.3475 \\
    \bottomrule
  \end{tabular}
  \label{table:retrieval}
\end{table}

\section{Recogniton Method}
\begin{figure*}[ht!]
  \centering
  \includegraphics[width=0.9\linewidth]{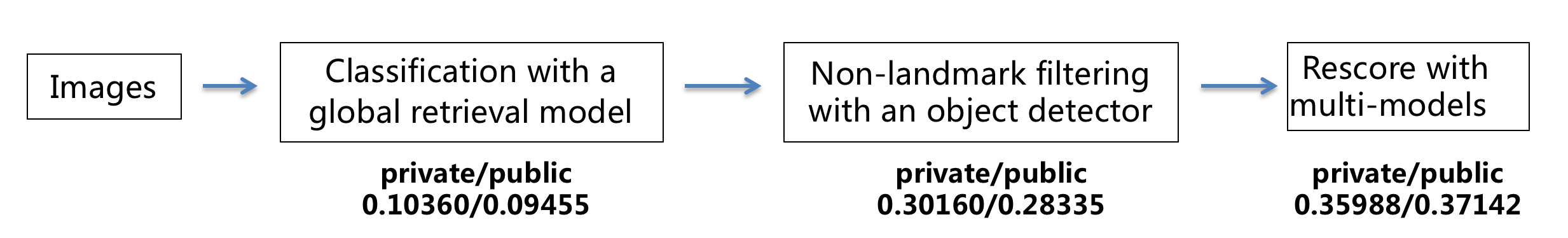}\\
  \caption{Our recognition pipeline to recognize landmark from test data}    
  \label{fig:rec_pipeline}
\end{figure*}

\subsection{Classification with a global retrieval model}
\begin{figure*}[ht!]
  \centering
  \includegraphics[width=0.9\linewidth]{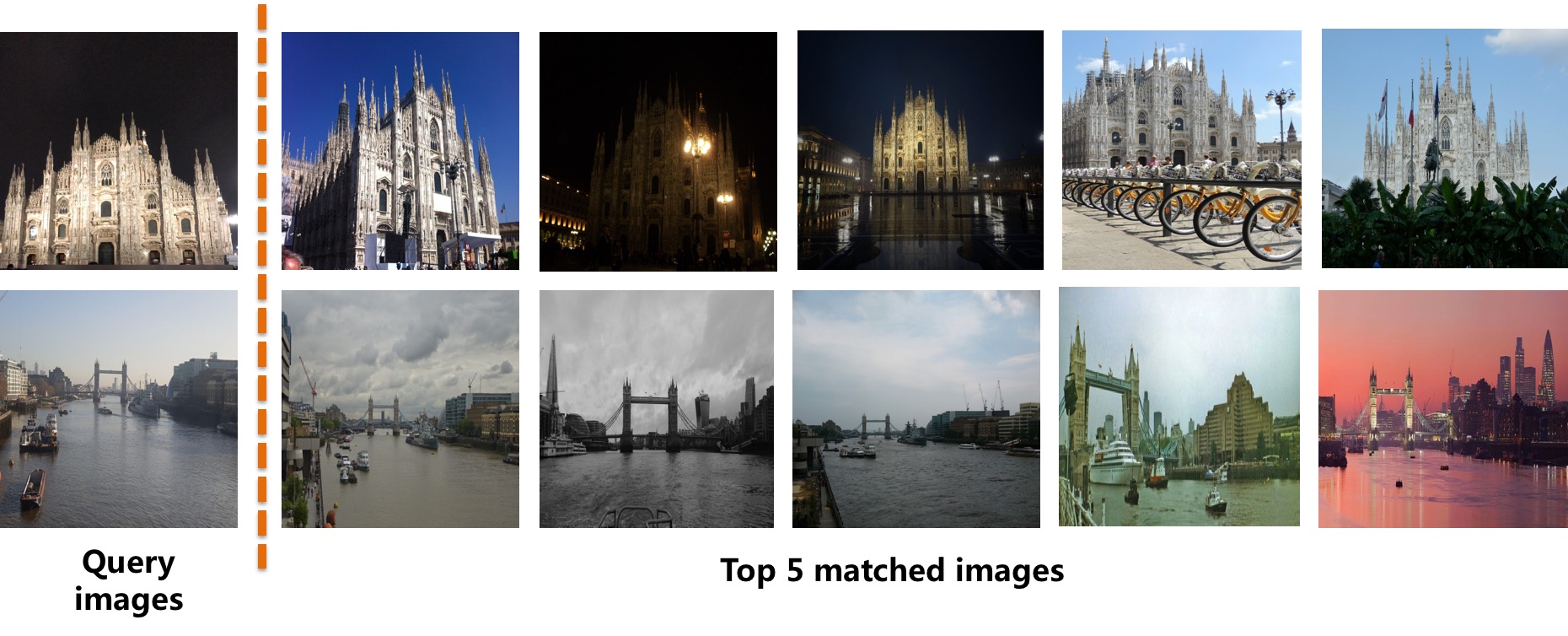}\\
  \caption{The match results of some challenging examples.}    
  \label{fig:rec_demo}
\end{figure*}

At first, we classify an image with k-nearest neighbors algorithm. The ResNet152 model which is mentioned in the retrieval task is used to extract the global feature of an image. Then match all test images (~120k) and all train images (~4.13M). For each test image, label an image by voting the top-5 matched images.The largest number of category is as the predicted label and the highest score is as the predicted score. Fig.~\ref{fig:rec_pipeline} shows the powerful of the global feature and some very challenging images can be correct recognized. However, since the retrieval task doesn't care non-landmark filtering, the scores of lots of non-landmarks are also very high and the GAP is only private/public 0.10360/0.09455.The match results of some challenging examples is shown in Fig.~\ref{fig:rec_demo}

\subsection{Non-landmark filtering with an object detector}
To filter the non-landmark images, we train a single object detector model based on Faster RCNN\cite{ren2015faster} with the open images dataset v4\footnote{\url{https://storage.googleapis.com/openimages/web/factsfigures_v4.html}} for object detection. The annotation files of the dataset cover the 600 boxable object classes, and span the 1,743,042 training images where the bounding boxes were annotated. We utilize Resnet50 as the backbone of above object detector and the mAP achieves about 0.55 of the public leaderboard of Google AI Open Images - Object Detection Track.  

To distinguish the non-landmark images, the 600 object classes are divided into three parts: landmark part, uncertain part and non-landmark part. The landmark part includes the following 5 classes: Building, Tower, Castle, Sculpture and Skyscraper. The uncertain part includes the following 7 classes: House, Tree, Palm tree, Watercraft, Aircraft, Swimming pool and Fountain. The other classes are considered in the non-landmark part. For a test image, if it exists at least one object in the landmark part, it is considered as a landmark image. If it exists one object in the non-landmark part, it is considered as a candidate non-landmark image. In order to keep more landmark images, two additional constraints are placed on the objects with non-landmark image. At first, the detector score of the object must be greater than 0.3. Secondly, the area ratio between the object bounding box and the whole image must be greater than 0.6. In this way, about 28k images from  the test images (~120k) are considered as the non-landmark images.  

To further filter the other non-landmark images,  we match all test images (~120k) and above filtered 28k images with the global retrieval feature. For a test image, when the minimal of top3 match scores above a certain threshold, this image is also deemed as a non-landmark image. In this way, about 64k the rest of images are filtered. When the above total 92k images are filtered, the performance improvement is quite obviously and the GAP achieves private/public 0.30160/0.28335.  

\subsection{Rerank with multi-models}
As everyone knows, since the GAP is related to the rank of all predictions, increasing highly credible landmark scores helps to improve the performance. So we grade and rescore the test images with multi-models, the above mentioned global retrieval model and a classification model. The classification model is based on ResNet152 and trained with  about 3M images. Those images are from the train images (4.13M) with above non-landmark filter method. Label smoothing is used in the training model, and the soft label parameter is set to 0.2. 

\begin{figure}
  \includegraphics[width=0.95\linewidth]{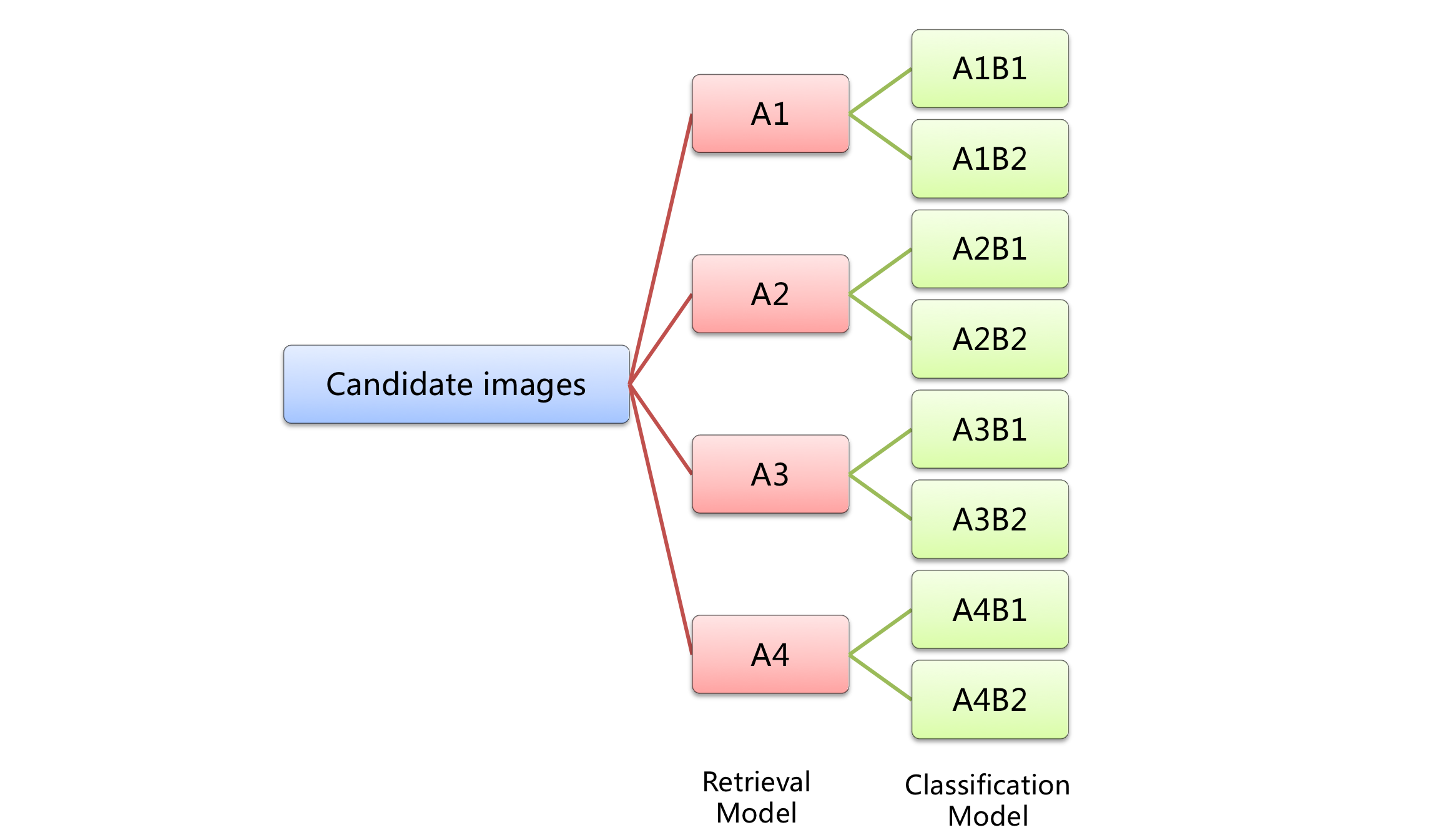}\\
  \caption{grade strategy used in recognition task} 
  \label{fig:rerank_level}
\end{figure}

Fig.~\ref{fig:rerank_level} shows our grade strategy. At first, we grade the landmark image with the retrieval model according to the top-5 matched images' labels and scores.  The largest number of category is as the predicted label and the highest score is as the predicted score.

The landmark images are divided into 4 parts: 

\begin{quotation}
\noindent
A1:  the number of the top-5 matched images' labels is not more than 2 and the minimum score of the predicted label is more than 0.9. 
\end{quotation}

\begin{quotation}
\noindent
A2: Similar as A1, the number of the top-5 matched images' labels is not more than 2, but this grade only asks that the maximum score of the predicted label is more than 0.85.
\end{quotation}

\begin{quotation}
\noindent
A3: the image does not satisfy A1, A2 or A4.
\end{quotation}

\begin{quotation}
\noindent
A4: the top-5 matched images' labels are all different.
\end{quotation}

When we grade above 4 parts and rescore the landmark image according to $A1 > A2 > A3 > A4$, the GAP achieves private/public 0.31340/0.29426. 

For each part, the landmark image is further graded with the classification model and is divided into 2 parts.
\begin{quotation}
\noindent
B1: the predicted label between the retrieval model and classification model are same.
\end{quotation}

\begin{quotation}
\noindent
B2: the image does not satisfy B1.
\end{quotation}

When we further grade and rescore the landmark image according to $A1B1 > A1B2 > A2B1 > A2B2 > A3B1 > A3B2 > A4B1 > A4B2$, the GAP achieves private/public 0.32574/0.30839.

In the first stage of competition, we discover an unbelievable trick: from the image set A1B1, the frequently occurring landmark categories set W can be found (we treated landmarks that appeared more than 5 times as the frequently  landmark). Then, top of the images in A1 and A2 which category in the set W and rescore these images according to the frequency of landmark category.  The GAP is improved from 0.14668 to 0.21657. Main reason for the promotion is that the distractors are seldom in the landmark images  with the object detector filter. So utilize the above trick can further dampen the  distractors. In the second stage of competition, adopt the same trick and rescore 431 images,  the GAP achieves private/public 0.35988/0.37142.  

After the competition, we adopt above trick not only in A1B1, but also in A1B2, A2B1, A2B2, A3B1, A3B2, then top the frequently occurring landmark images according to the grade and the frequency of landmark category. the GAP achieves private/public 0.37469/0.36365. Meanwhile, for each part, we top the images which categories are in the frequently occurring landmark categories set W of the first stage,  the GAP achieves private/public 0.38231/0.36805.

\section{Conclusion}
In this paper, we presented a large-scale image retrieval and recognition system by team imagesearch(GLRunner), including global feature, local feature, query extension, database augmentation, image rerank. The resulting fast research cycle allowed us to leverage several techniques that led to 2nd place in the Google Landmark Recognition 2019 and 2nd place in the Google Landmark Retrieval 2019 on kaggle.

{\small
\bibliographystyle{ieee_fullname}
\bibliography{egbib}
}

\end{document}